\documentclass[letterpaper]{article} 
\usepackage{aaai19}  
\usepackage{times}  
\usepackage{helvet}  
\usepackage{courier}  
\usepackage{url}  
\usepackage{graphicx}  
\usepackage{amsmath}
\usepackage{amssymb}
\usepackage{algorithm}
\usepackage{algorithmic}
\usepackage{makecell}
\usepackage{amsthm}
\usepackage{arydshln}
\usepackage{multirow}
\usepackage{makecell}

\title{Skeleton-to-Response: Dialogue Generation Guided by Retrieval Memory}

\author{Deng Cai\thanks{Work done while DC was interning at Tencent AI Lab.}\\\small The Chinese University of Hong Kong\\\small thisisjcykcd@gmail.com
	\And
	Yan Wang\\\small Tencent AI Lab\\\small brandenwang@tencent.com
	\And
	Victoria Bi\\\small Tencent AI Lab\\\small victoriabi@tencent.com
	\AND
	Zhaopeng Tu\\\small Tencent AI Lab\\\small zptu@tencent.com
	\And
	Xiaojiang Liu\\\small Tencent AI Lab\\\small kieranliu@tencent.com
	\And
	Wai Lam\\\small The Chinese University of Hong Kong\\\small wlam@se.cuhk.edu.hk
	\And
	Shuming Shi\\\small Tencent AI Lab\\\small shumingshi@tencent.com
}
\begin{document}
	\maketitle
	\begin{abstract}
		Traditional generative dialogue models generate responses solely from input queries. Such information is insufficient for generating a specific response since a certain query could be answered in multiple ways. Recently, researchers have attempted to fill the information gap by exploiting information retrieval techniques. For a given query, similar dialogues are retrieved from the entire training data and considered as an additional knowledge source. While the use of retrieval may harvest extensive information, the generative models could be overwhelmed, leading to unsatisfactory performance. In this paper, we propose a new framework which exploits retrieval results via a skeleton-to-response paradigm. At first, a skeleton is extracted from the retrieved dialogues. Then, both the generated skeleton and the original query are used for response generation via a novel response generator. Experimental results show that our approach significantly improves the informativeness of the generated responses.
	\end{abstract}
	\section{Introduction}
	This paper focuses on tackling the challenges to develop a chit-chat style dialogue system (also known as chatbot). Chit-chat style dialogue system aims at giving meaningful and coherent responses given a dialogue query in the open domain. Most modern chit-chat systems can be categorized into two categories, namely, information retrieval-based (IR) models and generative models.
	
	The IR-based models \cite{ji2014information,hu2014convolutional} directly copy an existing response from a training corpus when receiving a response request. Since the training corpus is usually collected from real-world conversations and possibly post-edited by a human, the retrieved responses are informative and grammatical. However, the performance of such systems drops when a given dialogue history is substantially different from those in the training corpus.
	
	The generative models \cite{shang-lu-li:2015:ACL-IJCNLP,vinyals2015neural,li-EtAl:2016:N16-11}, on the other hand, generate a new utterance from scratch. While those generative models have better generalization capacity in rare dialogue contexts, the generated responses tend to be universal and non-informative (e.g., ``I don't know'', ``I think so'' etc.) \cite{li-EtAl:2016:N16-11}. It is partly due to the diversity of possible responses to a single query (i.e., the \textit{one-to-many} problem). The dialogue query alone cannot decide a meaningful and specific response. Thus a well-trained model tends to generate the most frequent (safe but boring) responses instead.

	To summarize, IR-based models may give informative but inappropriate responses while generative models often do the opposite. It is desirable to combine both merits. \citeauthor{song2016two} (\citeyear{song2016two}) used an extra encoder for the retrieved response. The resulted dense representation, together with the original query, is used to feed the decoder in a standard \textsc{Seq2Seq} model \cite{bahdanau2014neural}. \citeauthor{weston2018retrieve} (\citeyear{weston2018retrieve}) used a single encoder that takes the concatenation of the original query and the retrieved as input. \citeauthor{wu2018response} (\citeyear{wu2018response}) noted that the retrieved information should be used in awareness of the context difference, and further proposed to construct an edit vector by explicitly encoding the lexical differences between the input query and the retrieved query.
	
	However, in our preliminary experiments, we found that the IR-guided models are inclined to degenerate into a copy mechanism, in which the generative models simply repeat the retrieved response without necessary modifications. Sharp performance drop is caused when the retrieved response is irrelevant to the input query. A possible reason is that both useful and useless information is mixed in the dense vector space, which is uninterpretable and uncontrollable.

	To address the above issue, we propose a new framework, skeleton-to-response, for response generation. Our motivations are two folds: (1) The guidance from IR results should only specify a response aspect or pattern, but leave the query-specific details to be elaborated by the generative model itself; (2) The retrieval results typically contain excessive information, such as inappropriate words or entities. It is necessary to filter out irrelevant words and derive a useful skeleton before use.
	
\begin{figure}[t]
		\centering
		\includegraphics[scale=0.7]{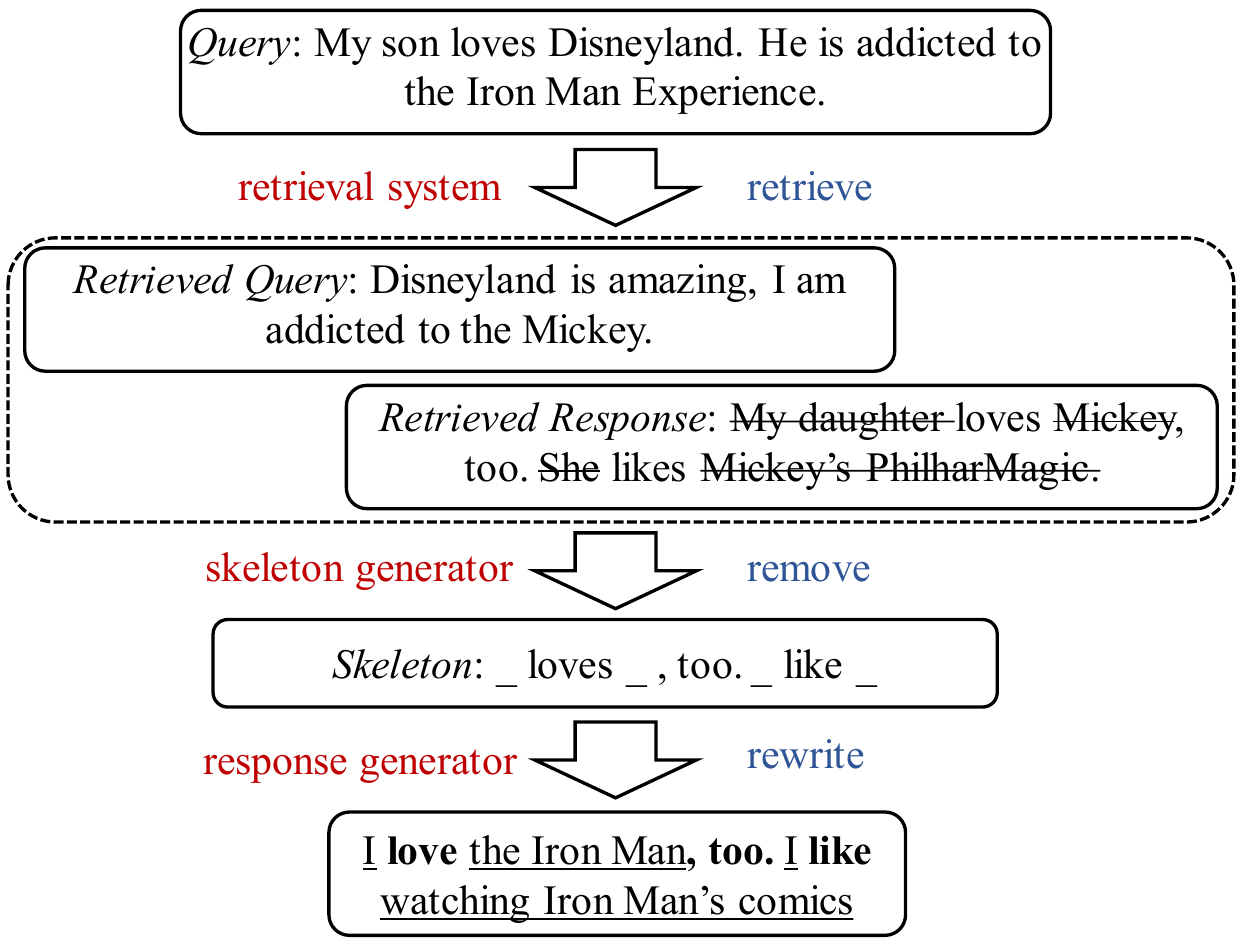}
		\caption{Our idea of leveraging the retrieved query-response pair. It first constructs a response skeleton by removing some words in the retrieved response, then a response is generated via rewriting based on the skeleton.}
		\label{flow}
	\end{figure}

	Our approach consists of two components: a skeleton generator and a response generator. The skeleton generator extracts a response skeleton by detecting and removing unwanted words in a retrieved response. The response generator is responsible for adding query-specific details to the generated skeleton for query-to-response generation. A dialogue example illustrating our idea is shown in Fig. \ref{flow}. Due to the discrete choice of skeleton words, the gradient in the training process is no longer differentiable from the response to the skeleton generator. Two techniques are proposed to solve this issue. The first technique is to employ the policy gradient method for rewarding the output of the skeleton generator based on the feedback from a pre-trained critic. An alternative technique is to solve both the skeleton generation and the response generation in a multi-task learning fashion.
	
	Our contributions are summarized as below: (1) We develop a novel framework to inject the power of IR results into generative response models by introducing the idea of skeleton generation; (2) Our approach generates response skeletons by detecting and removing unnecessary words, which facilitates the generation of specific responses while not spoiling the generalization ability of the underlying generative models; (3) Experimental results show that our approach significantly outperforms other compared methods, resulting in more informative and specific responses.
	\section{Models}
	\subsection{Overview}
	In this work, we propose to construct a response skeleton based on the result of IR systems for guiding the response generation. The skeleton-then-response paradigm helps reduce the output space of possible responses and provides useful elements missing in the current query.
	
	For each query $q \in Q$, a set of historical query-response pairs $R_q = \{(q'_i, r'_i)\}_{i=1}^{N}$ are retrieved by some IR techniques. We estimate the generation probability of a response $r$ conditioned on $q$ and $R_q$. The whole process is decomposed into two parts. First, we assume that there exists a probabilistic model $P_{\theta_{ske}}(t_i|q, q'_i, r'_i)$ mapping each $(q, q'_i, r'_i)$ to a response skeleton $t_i$. Basically, we mask some parts (ideally useless or unnecessary parts) of a retrieved response for producing a response skeleton. Armed with this skeleton, the final response is generated by revising the skeletons $T = \{t_i\}_{i=1}^{N}$ by $P_{\theta_{res}}(r | q, T)$. Our overall model consists of two components, namely, the skeleton generator and the response generator. These components are parameterized by the above two probabilistic models, denoted by $\theta_{ske}$ and $\theta_{res}$ respectively.
	
\begin{figure*}[t]
		\centering
		\includegraphics[scale=0.66]{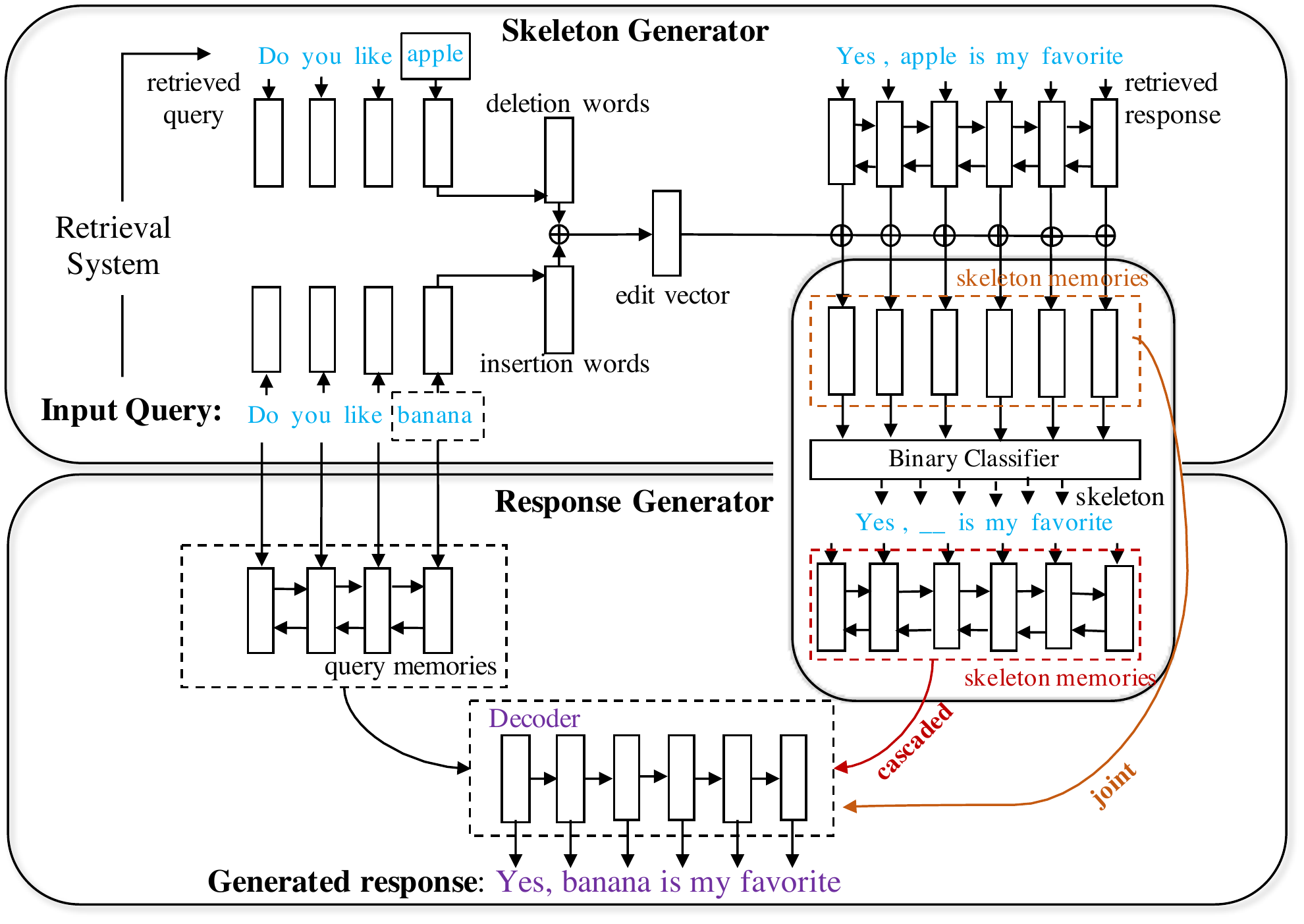}
		\caption{The architecture of our framework. Given a query ``Do you like banana", a similar historical query ``Do you like apple" is retrieved along with its response, i.e., ``Yes, apple is my favorite". Upper: The skeleton generator removes inappropriate words and extracts a response skeleton. Lower: The response generator generates a response based on both the skeleton and the query.}
		\label{arh}
	\end{figure*}

	For clarity, the proposed model is explained in detail under the default setting of $N= 1$ (i.e., $R_q = (q', r')$) in the following part of this section. It should be noted that our model is readily extended to incorporate multiple IR results. Fig. \ref{arh} depicts the architecture of our proposed framework.
	\subsection{Skeleton Generator}
	\label{skeleton}
	The skeleton generator transforms a retrieved response into a skeleton by explicitly removing inappropriate or useless information regarding the input query $q$. We consider this procedure as a series of word-level masking actions. Following \cite{wu2018response}, we first construct an edit vector by comparing the difference between the original query $q$ and the retrieved query $q'$. In \cite{wu2018response} the edit vector is used to guide the response generation directly. In our model, the edit vector is used to estimate the probability of being reserved or being masked for every word in a sentence. We define two word sets, namely \textit{insertion words} $I$ and \textit{deletion words} $D$. The insertion words include words that are in the original query $q$, but not in the retrieved query $q'$, while the deletion words do the opposite.
	
	The two bags of words highlight the changes in the dialogue context, corresponding to the changes in the response. The edit vector $z$ is thus defined as the concatenation of the representations of the two bags of words. We use the weighted sum of the word embeddings to get the dense representations of $I$ and $D$. The edit vector is computed as:
	\begin{equation}
	z= \sum_{w_1 \in I} \alpha_{w_1} \Phi(w_1 ) \oplus \sum_{w_2 \in D} \beta_{w_2} \Phi(w_2),
	\end{equation}
	where $\oplus$ is the concatenation operation. $\Phi$ maps a word to its corresponding embedding vector,  $\alpha_{w_1}$ and $\beta_{w_2}$ are the weights of an insertion word $w_1$ and a deletion word $w_2$ respectively. The weights of different words are derived by an attention mechanism \cite{luong-pham-manning:2015:EMNLP}. Formally, the retrieved response $r' = (r'_1, r'_2\ldots, r'_{|r'|})$ is processed by a bidirectional GRU network (biGRU). We denote the states of the biGRU (i.e. concatenation of forward and backward GRU states) as $(h_1, h_2, \ldots,h_{|r'|})$.
	The weight $\alpha_{w_1}$ is calculated by:
	\begin{equation}
	\alpha_{w_1} = \frac{\exp(s_{w_1})}{\sum_{w \in I }\exp(s_w)}, \nonumber
	\end{equation}
	\begin{equation}
	s_{w_1} ={v}_I^{\top} \tanh({W}_I[\Phi(w_1)\oplus h_{|r'|}]),
	\end{equation}
	where ${v}_I$ and ${W}_I$ are learnable parameters. The weight $\beta_{w_2}$ is obtained in a similar way with another set of parameters ${v}_D$ and ${W}_D$.
	
	After acquiring the edit vector, we transform the prototype response $r'$ to a skeleton $t$ by the following equations:
	\begin{equation}
	t = ( \phi( r'_1, h_1, z) , \phi( r'_2, h_2, z), \cdots, \phi( r'_{|r'|}, h_{|r'|}, z)), \nonumber
	\end{equation}
	\begin{equation}
	\phi(r'_i, h_i, z) =
	\begin{cases} <\text{blank}> &  \text{if $\hat{m}_i= 0$},\\
	r'_i & \text{else}
	\end{cases},
	\end{equation}
	where $\hat{m}_i$ is the indicator and equals 0 if $r'_i$ is replaced with a placeholder ``$<$blank$>$'' and 1 otherwise. The probability of $\hat{m}_i = 1$ is computed by
	\begin{equation}
	P( \hat{m}_i = 1) = \text{sigmoid}({W}_m[h_i \oplus z] + b_m).
	\end{equation}
	\subsection{Response Generator}
	The response generator can be implemented using most existing IR-augmented models \cite{song2016two,weston2018retrieve,pandey2018exemplar}, just by replacing the retrieved response input with the corresponding skeleton. We discuss our choices below.
	\paragraph{Encoders} Two separate bidirectional LSTM (biLSTM) networks are used to obtain the distributed representations of the query memories and the skeleton memories, respectively. For biLSTM, the concatenation of the forward and the backward hidden states at each token position is considered a memory slot, producing two memory pools: $\mathcal{M}_q =\{h_1, h_2, \ldots,h_{|q|}\}$ for the input query, and $\mathcal{M}_t =\{h'_1, h'_2, \ldots,h'_{|t|}\}$ for the skeleton.\footnote{Note the skeleton memory pool $\mathcal{M}_t$ could contain multiple response skeletons, further discussed in the experiment section.}
	\paragraph{Decoder}
	During the generation process, our decoder reads information from both the query and the skeleton using attention mechanism \cite{bahdanau2014neural,luong-pham-manning:2015:EMNLP}. To query the memory pools, the decoder uses the hidden state $s_t$ of itself as the searching key. The matching score function is implemented by bilinear functions:
	\begin{equation}
	\alpha(h_k, s_t) = {h_k}^{T}W_qs_t, \quad \beta(h'_k, s_t) = {h'_k}^{T}W_ts_t,
	\end{equation}
	where $W_q$ and $W_t$ are trainable parameters. A query context vector $c_t$  is then computed as a weighted sum of all memory slots in $\mathcal{M}_q$, where the weight for a memory slot $h_k$ is ${\exp(\alpha(h_k, s_t))}/({\sum_{i=1}^{|q|}\exp(\alpha(h_i, s_t))})$. A skeleton context vector $c'_t$ is computed in a similar spirit by using $\beta(h'_k, s_t)$'s.
	
	The probability of generating the next word $r_t$ is then jointly determined by the decoder's state $s_t$, the query context $c_t$ and the skeleton context $c'_t$. We first fuse the information of $s_t$ and $c_t$ by a linear transformation. For $c'_t$, a gating mechanism is additionally introduced to control the information flow from skeleton memories. Formally, the probability of the next token $r_t$ is estimated by $y_t$ followed by a softmax function over the vocabulary:
	\begin{equation}
	y_t = (W_c[s_t \oplus c_t])\cdot g_t + c'_t \cdot (1-g_t),
	\end{equation}
	where $g_t = f_g(s_t,c_t, c'_t)$ is implemented by a single layer neural network with sigmoid output layer.
	\section{Learning}
	Given that our skeleton generator performs non-differentiable hard masking, the overall model cannot be trained end-to-end using the standard maximum likelihood estimate (MLE). A possible solution that circumvents this problem is to treat the skeleton generation and the response generation as two parallel tasks and solve them jointly in a multi-task learning fashion. An alternative is to bridge the skeleton generator and the final response output using reinforcement learning (RL) methods, which can exclusively inform the skeleton generator with the ultimate goal. The latter option is referred as \textit{cascaded integration} while the former is called \textit{joint integration}.
	
	Recall that we have formulated the skeleton generation as a series of binary classifications. Nevertheless, most of the dialogue datasets are end-to-end query-response pairs without explicit skeletons. Hence, we propose to construct \textit{proxy skeleton}s to facilitate the training.
	
	\textbf{Definition 1} Proxy Skeleton: \textit{Given a training quadruplet $(q, q', r, r')$ and a stop word list $S$, the proxy skeleton for $r$ is generated by replacing some tokens in $r'$ with a placeholder ``$<$blank$>$''. A token $r'_i$ is kept if and only if it meets the following conditions \\
		\indent 1. $r'_i\notin S$ \\
		\indent 2. $r'_i$ is a part of the longest common sub-sequence (LCS) \cite{wagner1974string} of $r$ and $r'$.}
	\begin{algorithm}[t]
		\caption{Proxy Skeleton Construction}
		\begin{algorithmic}[1]
			\REQUIRE a training quadruplet $(q, q', r, r')$, stop word list $S$
			\ENSURE the proxy skeleton $t$, the proxy labels $m$.
			\STATE $r^*, {r'}^*$ $\leftarrow$ remove the stop words in $r$ and $r'$
			\STATE $x  \leftarrow$ LongestCommonSubsequence$(r^*, {r'}^*)$
			\FOR {$i=1$ to $|r'|$} 
			\STATE $m_i  \leftarrow 1$ if $(r'_i \in x$ and $r'_i \notin S)$ else $0$
			\STATE $t_i  \leftarrow r'_i$ if $(m_i = 1)$ else ``$<$blank$>$''
			\ENDFOR
			\RETURN $t, m$
		\end{algorithmic}
		\label{algo}
	\end{algorithm}
	
	The detailed construction process is given in Algorithm \ref{algo}. The proxy skeletons are used in different manners according to the integration method, which we will introduce below.
	\subsubsection{Joint Integration} To avoid breaking the differentiable computation, we connect the skeleton generator and the response generator via shared network architectures rather than by passing the discrete skeletons. Concretely, the last hidden states in our skeleton generator (i.e, the hidden states that are utilized to make the masking decisions) are directly used as the skeleton memories in response generation. The skeleton generation and response generation are considered as two tasks. For skeleton generation, the object is to maximize the log likelihood of the proxy skeleton labels:
	\begin{equation}
	L(\theta_{ske}) = \sum_{i=1}^{|r'|}\log P(m_i |  q, q', r'),\label{mle_ske}
	\end{equation}
	while for response generation, it is trained to maximize the following log likelihood:
	\begin{equation}
	L(\theta_{res}) = \sum_{i=1}^{|r|} \log P(r_i | r_{1:i-1}, q, t). \label{mle_res}
	\end{equation}
	The joint network is then trained to maximize two parts of log likelihood:
	\begin{equation}
	L(\theta_{res} \cup \theta_{ske}) = L(\theta_{res}) + \eta L(\theta_{ske}) ,
	\end{equation}
	where $\eta$ is a harmonic weight, and it is set as $1.0$ in our experiments.
	\subsubsection{Cascaded Integration} Policy gradient methods \cite{williams1992simple} can be applied to optimize the full model while keeping it running as cascaded process. We regard the skeleton generator as the first RL agent, and the response generator as the second one. The final output generated by the pipeline process and the intermediate skeleton are denoted by $\hat{r}$ and $\hat{t}$ respectively. Given the original query $q$ and the generated response $\hat{r}$, a reward $R(q, \hat{r})$ for generating $\hat{r}$ is calculated. All network parameters are then optimized to maximize the expected reward by the policy gradient. According to the policy gradient theorem \cite{williams1992simple}, the gradient for the first agent is
	\begin{equation}
	\nabla_{\theta_{ske}} J(\theta_{ske}) = \mathbb{E} [ R \cdot \nabla \log(P(\hat{t}| q, q', r'))],
	\end{equation}
	and the gradient for the second agent is 
	\begin{equation}
	\nabla_{\theta_{res}} J(\theta_{res}) = \mathbb{E} [ R \cdot \nabla \log(P(\hat{r}| q, \hat{t}))].
	\end{equation}
	
	The reward function $R$ should convey both the naturalness of the generated response and its relevance to the given query $q$. A pre-trained critic is utilized to make the judgment. Inspired by comparative adversarial learning in \cite{li2018generating}, we design the critic as a classifier that receives four inputs every time: the query $q$, a human-written response $r$, a machine-generated response $\hat{r}$ and a random response $\overline{r}$ (yet written by human). The critic is trained to correctly pick the human-written response $r$ among others. Formally, the following objective is maximized:
	\begin{equation}
	\log D(r| q,\hat{r},\overline{r}, r ) = \log\frac{\exp({h_r}^\text{T}M_D h_q)}{\sum_{x \in \{\hat{r},\overline{r}, r \}}\exp({h_x}^{\text{T}}M_D h_q)},
	\end{equation}
	where $h_x$ is a vector representation of $x$, produced by a bidirectional LSTM (the last hidden state), and $M_D$ is a trainable matrix.\footnote{Note the classifier could be fine-tuned with the training of our generators, which falls into the adversarial learning setting \cite{goodfellow2014generative}.} The reward function of $\hat{r}$ is defined as:
	\begin{equation}
	R(q, {\hat{r}})= \log\frac{\exp({h_{\hat{r}}}^{\text{T}}M_D h_q)}{\sum_{x \in \{\hat{r},\overline{r}, r \}}\exp({h_x}^{\text{T}}M_D h_q)}.
	\end{equation}
	
	However, when randomly initialized, the skeleton generator and the response generator transmit noisy signals to each other, which leads to sub-optimal policies. We hence propose pre-training each component using Equation (\ref{mle_ske}) and (\ref{mle_res}) sequentially.
	\section{Related Work}
	\paragraph{Multi-source Dialogue Generation} Chit-chat style dialogue system dates back to ELIZA \cite{weizenbaum1966eliza}. Early work uses handcrafted rules, while modern systems usually use data-driven approaches, e.g., information retrieval techniques. Recently, end-to-end neural approaches \cite{vinyals2015neural,serban2016building,li-EtAl:2016:N16-11,sordoni2015neural} have attracted increasing interest. For those generative models, a notorious problem is the ``safe response" problem: the generated responses are dull and generic, which may attribute to the lack of sufficient input information. The query alone cannot specify an informative response. To mitigate the issue, many research efforts have been paid to introducing other information source, such as unsupervised latent variable \cite{serban2017hierarchical,zhao2018unsupervised,cao2017latent,shen2017conditional},  discourse-level variations \cite{zhao2017learning}, topic information \cite{xing2017topic}, speaker personality  \cite{li2016persona} and knowledge base \cite{ghazvininejad2017knowledge,zhou2018commonsense}. Our work follows the similar motivation and uses the output of IR systems as the additional knowledge source.
	\paragraph{Combination of IR and Generative models} To combine IR and generative models, early work \cite{qiu-EtAl:2017:Short} tried to re-rank the output from both models. However, the performance of such models is limited by the capacity of individual methods. Most related to our work, \cite{song2016two,weston2018retrieve} and \cite{wu2018response} encoded the retrieved result into distributed representation and used it as the additional conditionals along with the standard query representation. While the former two only used the target side of the retrieved pairs, the latter took advantages of both sides. In a closed domain conversation setting, \cite{pandey2018exemplar} further proposed to weight different training instances by context similarity. Our model differs from them in that we take an extra intermediate step for skeleton generation to filter the retrieval information before use, which shows the effectiveness in avoiding the erroneous copy in our experiments. 
	\paragraph{Multi-step Language Generation}
	Our work is also inspired by recent successes of decomposing an end-to-end language generation task into several sequential sub-tasks. For document summarization, \citeauthor{chen2018fast}   (\citeyear{chen2018fast}) first select salient sentences and then rewrite them in parallel. For sentiment-to-sentiment translation, \citeauthor{unpaired-sentiment-translation} (\citeyear{unpaired-sentiment-translation}) first use a neutralization module to remove emotional words and then add sentiment to the neutralized content. Not only does their decomposition improve the overall performance, but also makes the whole generation process more interpretable. Our skeleton-to-response framework also sheds some light on the use of retrieval memories.
	\section{Experiments}
	\subsection{Data} We use the preprocessed data in \cite{wu2018response} as our test bed. The total dataset consists of about 20 million single-turn query-response pairs collected from Douban Group\footnote{https://www.douban.com/group}. Since similar contexts may correspond to totally different responses, the training quadruples $(q, r, q', r')$ for IR-augmented models are constructed based on response similarity. All response are indexed by Lucene.\footnote{https://lucene.apache.org/core/} For each $(q, r)$ pair, top 30 similar responses with their corresponding contexts are retrieved $\{(q'_i, r'_i) \}_{i=1}^{30}$. However, only those satisfying $0.3 \le Jaccard(r, r'_i) \le 0.7$ are leveraged for training, where $Jaccard$ measures the Jaccard distance. The reason for the data filter is that nearly identical responses drive the model to do simple copy while distantly different responses make the model ignore the retrieval input. About 42 million quadruples are obtained afterward. 
	
	For computational efficiency, we randomly sample 5 million quadruples as training data for all experiments. The test set consists of 1,000 randomly selected queries that are not in our training data.\footnote{Note the retrieval results for test data are based on query similarity, and no data filter is adopted.} For a fair comparison, when training a generative model without the help of IR, the quadruples are split to pairs.
	\subsection{Model Details}
	We implement the skeleton generator based on a bidirectional recurrent neural network with 500 LSTM units. We concatenate the hidden states from both directions. The word embedding size is set to 300. For the response generator, the encoder for queries, the encoder for skeletons and the decoder are three two-layer recurrent neural networks with 500 LSTM units where both encoders are bidirectional. We use dropout \cite{srivastava2014dropout} to alleviate overfitting. The dropout rate is set to 0.3 across different layers. The same architecture for the encoders and the decoder is shared across the following baseline models, if applicable.
	\subsection{Compared Methods}
	\begin{itemize}
		\item \textbf{Seq2Seq} the standard attention-based RNN encoder-decoder model \cite{bahdanau2014neural}.
		\item \textbf{MMI} \textsc{Seq2Seq} with Maximum Mutual Information (MMI) objective in decoding \cite{li-EtAl:2016:N16-11}. In practice, an inverse (response-to-query) \textsc{Seq2Seq} model is used to rerank the $N$-best hypothesizes from the standard \textsc{Seq2Seq} model ($N$ equals 100 in our experiments).
		\item \textbf{EditVec} the model proposed in \cite{wu2018response}, where the edit vector $z$ is used directly at each decoding step by concatenating it to the word embeddings.
		\item \textbf{IR} the Lucene system is also used a benchmark.\footnote{Note IR selects response candidates from the entire data collection, not restricted to the filtered one.}
		\item \textbf{IR+rerank} rerank the results of \textbf{IR} by \textbf{MMI}.
	\end{itemize}
	Besides, We use  \textbf{JNT} to denote our model with joint integration, and \textbf{CAS} for our model with cascaded integration. To validate the usefulness of the proposed skeletons. We design a response generator that takes an intact retrieval response as its skeleton input (i.e., to completely skip the skeleton generation step), denoted by \textbf{SKP}.\footnote{There are some other IR-augmented models using standard \textsc{seq2seq} models as SKP. \citeauthor{weston2018retrieve} (\citeyear{weston2018retrieve})used a rule to select either the generated response or the retrieved response as output, while we would like to focus on improving the quality of generated responses. \citeauthor{pandey2018exemplar} (\citeyear{pandey2018exemplar}) concentrated on closed domain conversations, their hierarchical encoder is not suitable for our open domain setting. We thus omit the empirical comparison with them.}
	\begin{table}[t]
		\centering
		\begin{tabular}{c|ccc}
			\hline
			model & human score & dist-1 & dist-2\\
			\hline
			IR & 2.093  & \textbf{0.238} & \textbf{0.723} \\
			IR+rerank & 2.520 & 0.208 & 0.586\\
			\hdashline
			Seq2Seq & 2.433  &0.156 & 0.336\\
			MMI &2.554  & 0.170 &0.464 \\
			EditVec & 2.588$^\dagger$&0.154 &0.394 \\
			SKP & 2.581 & 0.152&0.406 \\
			\hdashline
			JNT & 2.612$^\dagger$ & 0.147&0.377 \\
			CAS & \textbf{2.747}&0.156 &0.411 \\
			\hline
		\end{tabular}
		\caption{Response performance of different models. Sign tests on human score show that the CAS is significantly better than all other methods with p-value $<0.05$, and the p-value $<0.01$ except for those marked by $\dagger$. }
		\label{res}
	\end{table}
	\begin{figure}[t]
		\centering
		\includegraphics[scale=0.5]{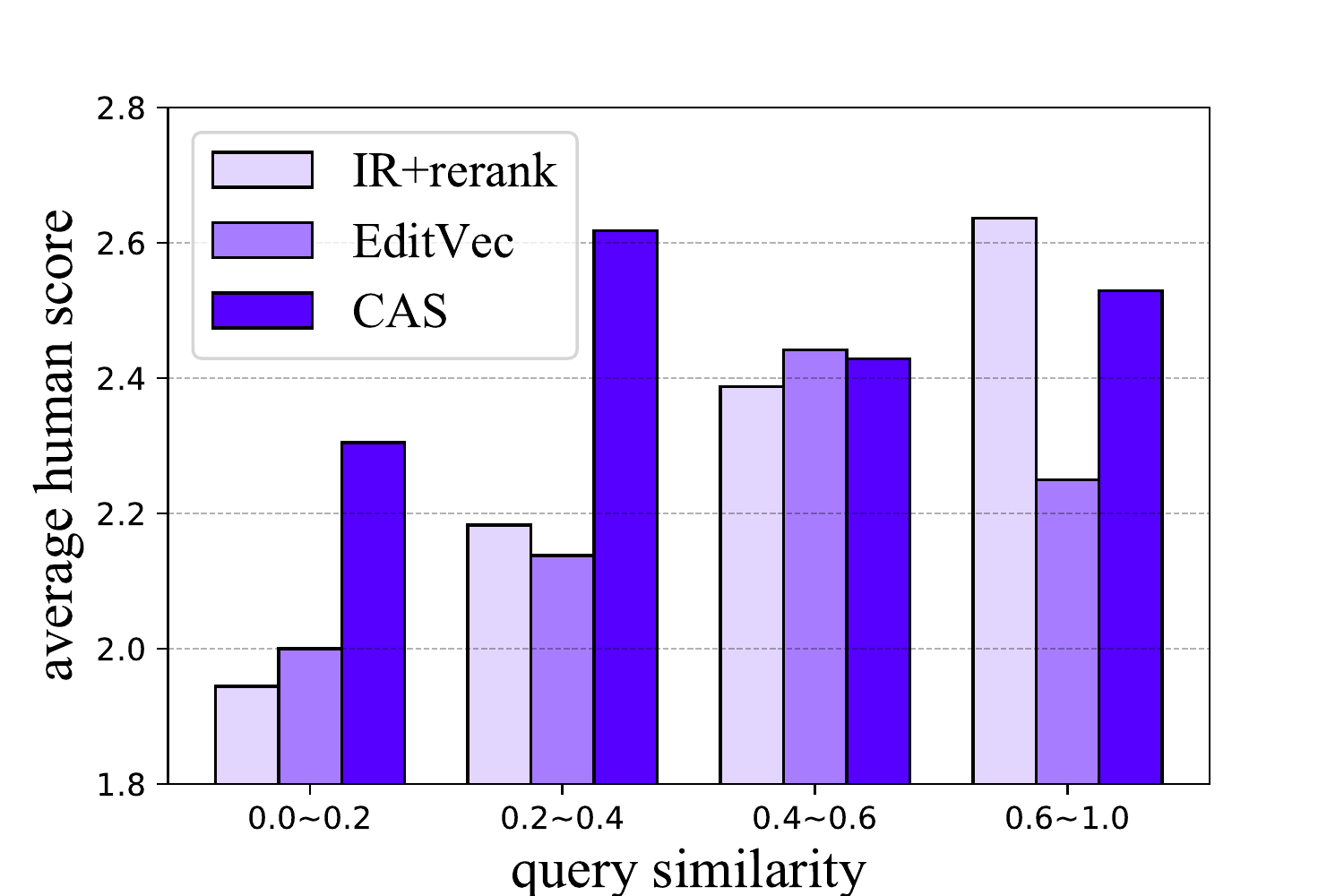}
		\caption{Response quality v.s. query similarity.\footnotemark}
		\label{quality-sim1}
	\end{figure}
	\begin{table}[t]
		\centering
		\begin{tabular}{c|ccccc}
			\hline
			model &P &R& F$_1$ & Acc.\\
			\hline
			JNT& 0.32&0.61&0.42&0.60\\
			CAS& 0.50 &0.86 & 0.63&0.76\\        
			\hline
		\end{tabular}
		\caption{Performance of skeleton generator.}
		\label{ske}
	\end{table}
	\subsection{Evaluation Metrics}
	Our method is designed to promote the informativeness of the generative model and alleviate the inappropriateness problem of the retrieval model. To measure the performance effectively, we use human evaluation along with two automatic evaluation metrics.
	\begin{itemize} 
		\item \textbf{Human evaluation}
		We asked three experienced annotators to score the group of responses (the best output of each model) for 300 test queries. The responses are rated on a five-point scale. A response should be scored 1 if it can hardly be considered a valid response, 3 if it is a valid but not informative response, 5 if it is an informative response, which can deepen the discussion of the current topic or lead to a new topic. 2 and 4 are for decision dilemmas.
		\item \textbf{dist-1 \& dist-2}  It is defined as the number of unique uni-grams (\textbf{dist-1}) or bi-grams (\textbf{dist-2}) dividing by the total number of tokens, measuring the diversity of the generated responses \cite{li-EtAl:2016:N16-11}. Note the two metrics do not necessarily reflect the response quality as the target queries are not taken into consideration.
	\end{itemize}
	\subsection{Response Generation Results}
	The results are depicted in Table \ref{res}. Overall, both of our models surpass all other methods, and our cascaded model (CAS) gives the best performance according to human evaluation. The contrast with the SKP model illustrates that the use of skeletons brings a significant performance gain.
	
	According to the dist-1\&2 metrics, the generative models achieve significantly better diversity by the use of retrieval results. The retrieval method yields the highest diversity, which is consistent with our intuition that the retrieval responses typically contain a large amount of information though they are not necessarily appropriate. The model of MMI also gives strong diversity, yet we find that it tends to merely repeat the words in queries. By removing the words in queries, the dist-2 of MMI and CAS become 0.710 and 0.751 respectively. It indicates our models are better at generating new words.
	
	To further reveal the source of performance gain, we study the relation between response quality and query similarity (measured by the Jaccard similarity between the input query and the retrieved query). Our best model (CAS) is compared with the strong IR system (IR-rerank) and the previous state-of-the-art (EditVec) in Fig. \ref{quality-sim1}. The CAS model significantly boosts the performance when query similarity is relatively low, which indicates that introducing skeletons can alleviate erroneous copy and keep a strong generalization ability of the underlying generative model.\footnotetext{We merge the ranges $[0.6, 0.8]$ and $[0.8, 1.0]$ due to the sparsity of highly similar pairs.}
	\begin{table*}[t]
		\centering
		\includegraphics[scale=0.82]{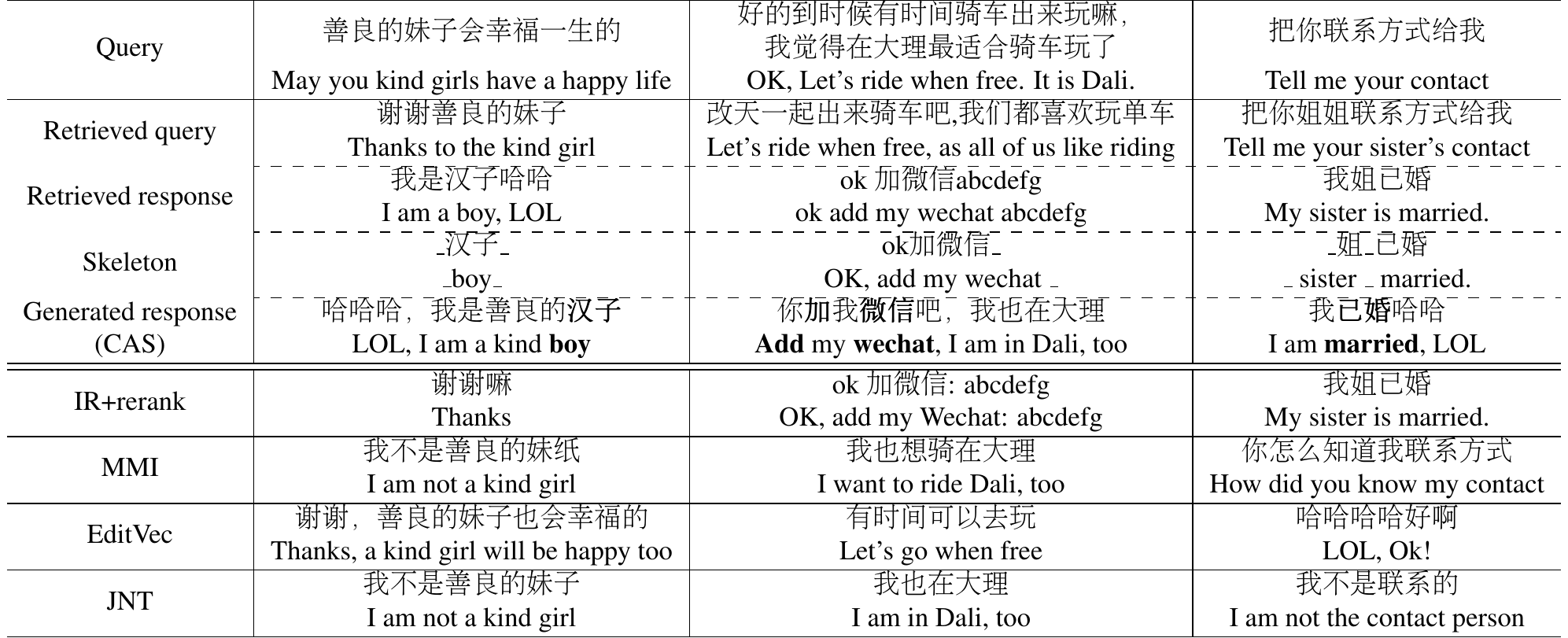}
		\caption{Upper: Skeleton-to-response examples of the CAS model. Lower: Responses from different models are for comparison.}
		\label{case-ske}
	\end{table*}
	\subsection{More Analysis of Our Framework}
	Here, we present further discussions and empirical analysis of our framework.
	\paragraph{Generated Skeletons}
	Although generating skeletons is not our primary goal, it is interesting to assess the skeleton generation. The word-level precision (P), recall (R), F$_1$ score (F$_1$) and accuracy (Acc.) of the well-trained skeleton generators are reported in Table \ref{ske}, taking the proxy skeletons as golden references. 
	
\begin{figure}[t]
		\centering
		\includegraphics[scale=0.5]{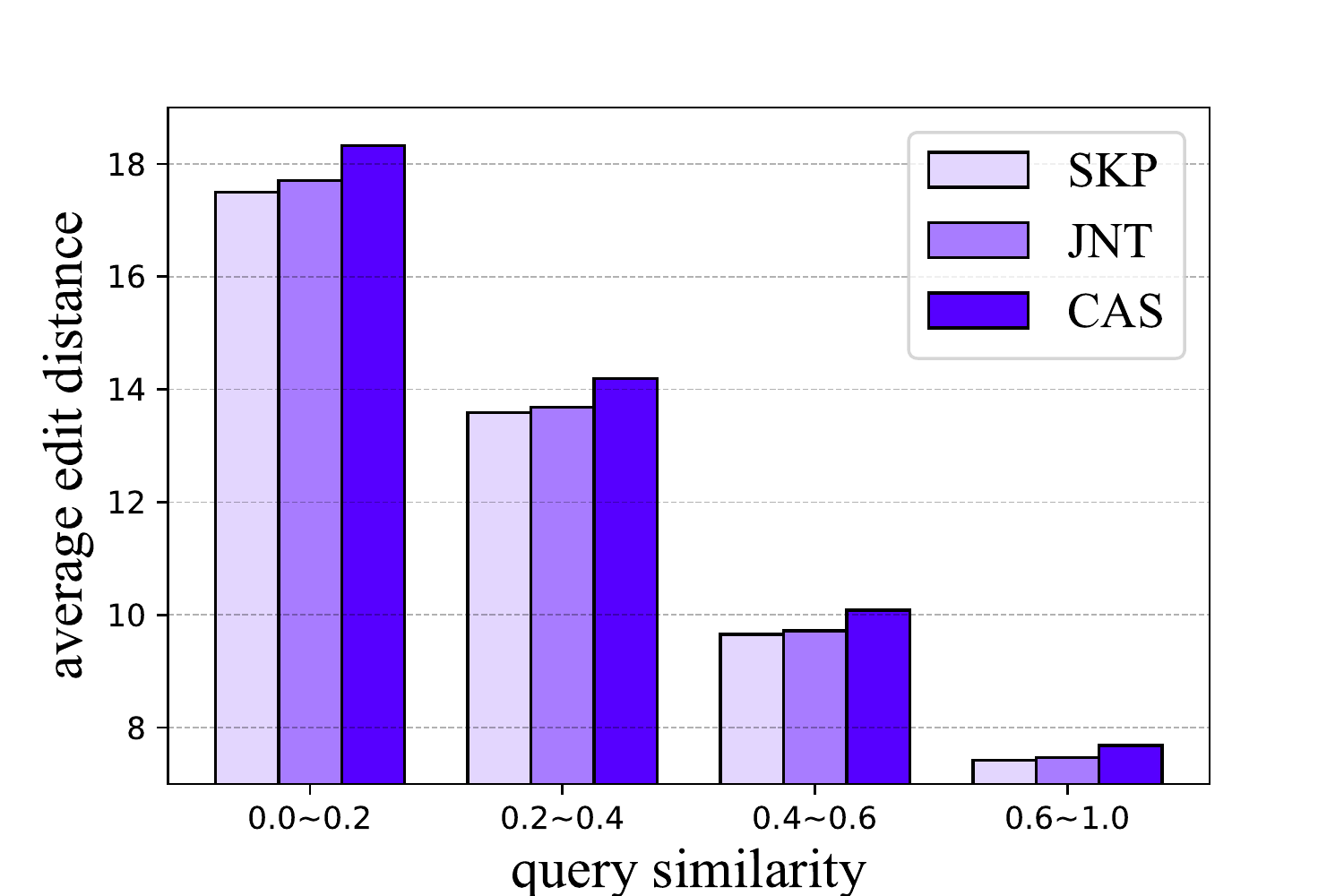}
		\caption{Changes between retrieved and generated responses v.s. query similarity.}
		\label{change-sim}
	\end{figure}

	Table \ref{case-ske} shows some skeleton-to-response examples of the CAS model and a case study among different models. In the leftmost example in Table \ref{case-ske}, the MMI and the EditVec simply repeat the query while the retrieved response is weakly related to the query. Our CAS model extracts a useful word 'boy' from the retrieved response and generates a more interesting response. In the middle example, the MMI response makes less sense, and some private information is included in the retrieved response. Our CAS model removes the privacy without the loss of informativeness, while the outputs by other models are less informative. The rightmost case shows that our response generator is able to recover the possible mistakes made by the skeleton generator.
	\paragraph{Retrieved Response v.s. Generated Response}
	To measure the extent that the generative models are copying the retrieval, we compute the edit distances between generated responses and retrieved responses. As shown in Fig. \ref{change-sim}, in the comparison between the SKP and other models, the use of skeletons makes the generated response deviate more from its prototype response. Ideally, when the retrieved context is very similar to the input query, the changes between the generated response and the prototype response should be minor. Conversely, the changes should be drastic. Fig. \ref{change-sim} also shows that our models can learn this intuition.
	\paragraph{Single v.s. Multiple Retrieval Pair(s)} For a given query $q$, the retrieval pair set $R_q$ could contain multiple query-response pairs. We investigate two ways of using it under the CAS setting.
	\begin{itemize}
		\item \textbf{Single} For each query-response pair $(q'_i, r'_i) \in R_q$, a response $\hat{r}_i$ is generated solely based on $q$, and $(q'_i, r'_i)$. The resulted responses are re-ranked by generation probability.
		\item \textbf{Multiple} The whole retrieval set $R_q$ is used in a single run. Multiple skeletons are generated and concatenated in the response generation stage.
	\end{itemize}
	The results are shown in Table \ref{single-multiple}. We attribute the failure of \textbf{Multiple} to the huge variety of the retrieved responses. The response generator receives many heterogeneous skeletons, yet it has no idea which to use. It remains an open question on how to effectively use multiple retrieval pairs for generating one single response, and we leave it for future work.
	\begin{table}[t]
		\centering
		\begin{tabular}{c|ccc}
			\hline
			setting & human score & dist-1 & dist-2 \\ 
			\hline
			Single& 2.747 & 0.156 & 0.411\\
			Multiple &1.976 & 0.178 & 0.414\\
			\hline
		\end{tabular}
		\caption{Comparison of the usages of the retrieval set.}
		\label{single-multiple}
	\end{table}
	\section{Conclusion}
	In this paper, we proposed a new methodology to enhance generative models with information retrieval technologies for dialogue response generation. Given a dialogue context, our methods generate a skeleton based on historical responses that respond to a similar context. The skeleton serves as an additional knowledge source that helps specify the response direction and complement the response content. Experiments on real world data validated the effectiveness of our method for more informative and appropriate responses.
	\bibliography{aaai19}

\begin{thebibliography}{}

\bibitem[\protect\citeauthoryear{Bahdanau, Cho, and
  Bengio}{2014}]{bahdanau2014neural}
Bahdanau, D.; Cho, K.; and Bengio, Y.
\newblock 2014.
\newblock Neural machine translation by jointly learning to align and
  translate.
\newblock In {\em ICLR}.

\bibitem[\protect\citeauthoryear{Cao and Clark}{2017}]{cao2017latent}
Cao, K., and Clark, S.
\newblock 2017.
\newblock Latent variable dialogue models and their diversity.
\newblock In {\em EACL},  182--187.

\bibitem[\protect\citeauthoryear{Chen and Bansal}{2018}]{chen2018fast}
Chen, Y.-C., and Bansal, M.
\newblock 2018.
\newblock Fast abstractive summarization with reinforce-selected sentence
  rewriting.
\newblock In {\em ACL}.

\bibitem[\protect\citeauthoryear{Ghazvininejad \bgroup et al\mbox.\egroup
  }{2018}]{ghazvininejad2017knowledge}
Ghazvininejad, M.; Brockett, C.; Chang, M.-W.; Dolan, B.; Gao, J.; Yih, W.-t.;
  and Galley, M.
\newblock 2018.
\newblock A knowledge-grounded neural conversation model.
\newblock In {\em AAAI},  5110--5117.

\bibitem[\protect\citeauthoryear{Hu \bgroup et al\mbox.\egroup
  }{2014}]{hu2014convolutional}
Hu, B.; Lu, Z.; Li, H.; and Chen, Q.
\newblock 2014.
\newblock Convolutional neural network architectures for matching natural
  language sentences.
\newblock In {\em NIPS},  2042--2050.

\bibitem[\protect\citeauthoryear{Ji, Lu, and Li}{2014}]{ji2014information}
Ji, Z.; Lu, Z.; and Li, H.
\newblock 2014.
\newblock An information retrieval approach to short text conversation.
\newblock {\em arXiv preprint arXiv:1408.6988}.

\bibitem[\protect\citeauthoryear{Li \bgroup et al\mbox.\egroup
  }{2016a}]{li-EtAl:2016:N16-11}
Li, J.; Galley, M.; Brockett, C.; Gao, J.; and Dolan, B.
\newblock 2016a.
\newblock A diversity-promoting objective function for neural conversation
  models.
\newblock In {\em NAACL},  110--119.

\bibitem[\protect\citeauthoryear{Li \bgroup et al\mbox.\egroup
  }{2016b}]{li2016persona}
Li, J.; Galley, M.; Brockett, C.; Spithourakis, G.~P.; Gao, J.; and Dolan, B.
\newblock 2016b.
\newblock A persona-based neural conversation model.
\newblock In {\em ACL},  994--1003.

\bibitem[\protect\citeauthoryear{Li \bgroup et al\mbox.\egroup
  }{2018}]{li2018generating}
Li, D.; He, X.; Huang, Q.; Sun, M.-T.; and Zhang, L.
\newblock 2018.
\newblock Generating diverse and accurate visual captions by comparative
  adversarial learning.
\newblock {\em arXiv preprint arXiv:1804.00861}.

\bibitem[\protect\citeauthoryear{Luong, Pham, and
  Manning}{2015}]{luong-pham-manning:2015:EMNLP}
Luong, T.; Pham, H.; and Manning, C.~D.
\newblock 2015.
\newblock Effective approaches to attention-based neural machine translation.
\newblock In {\em EMNLP},  1412--1421.

\bibitem[\protect\citeauthoryear{Pandey \bgroup et al\mbox.\egroup
  }{2018}]{pandey2018exemplar}
Pandey, G.; Contractor, D.; Kumar, V.; and Joshi, S.
\newblock 2018.
\newblock Exemplar encoder-decoder for neural conversation generation.
\newblock In {\em ACL},  1329--1338.

\bibitem[\protect\citeauthoryear{Qiu \bgroup et al\mbox.\egroup
  }{2017}]{qiu-EtAl:2017:Short}
Qiu, M.; Li, F.-L.; Wang, S.; Gao, X.; Chen, Y.; Zhao, W.; Chen, H.; Huang, J.;
  and Chu, W.
\newblock 2017.
\newblock Alime chat: A sequence to sequence and rerank based chatbot engine.
\newblock In {\em ACL},  498--503.

\bibitem[\protect\citeauthoryear{Serban \bgroup et al\mbox.\egroup
  }{2016}]{serban2016building}
Serban, I.~V.; Sordoni, A.; Bengio, Y.; Courville, A.~C.; and Pineau, J.
\newblock 2016.
\newblock Building end-to-end dialogue systems using generative hierarchical
  neural network models.
\newblock In {\em AAAI}, volume~16,  3776--3784.

\bibitem[\protect\citeauthoryear{Serban \bgroup et al\mbox.\egroup
  }{2017}]{serban2017hierarchical}
Serban, I.~V.; Sordoni, A.; Lowe, R.; Charlin, L.; Pineau, J.; Courville,
  A.~C.; and Bengio, Y.
\newblock 2017.
\newblock A hierarchical latent variable encoder-decoder model for generating
  dialogues.
\newblock In {\em AAAI},  3295--3301.

\bibitem[\protect\citeauthoryear{Shang, Lu, and
  Li}{2015}]{shang-lu-li:2015:ACL-IJCNLP}
Shang, L.; Lu, Z.; and Li, H.
\newblock 2015.
\newblock Neural responding machine for short-text conversation.
\newblock In {\em ACL},  1577--1586.

\bibitem[\protect\citeauthoryear{Shen \bgroup et al\mbox.\egroup
  }{2017}]{shen2017conditional}
Shen, X.; Su, H.; Li, Y.; Li, W.; Niu, S.; Zhao, Y.; Aizawa, A.; and Long, G.
\newblock 2017.
\newblock A conditional variational framework for dialog generation.
\newblock In {\em ACL},  504--509.

\bibitem[\protect\citeauthoryear{Song \bgroup et al\mbox.\egroup
  }{2016}]{song2016two}
Song, Y.; Yan, R.; Li, X.; Zhao, D.; and Zhang, M.
\newblock 2016.
\newblock Two are better than one: An ensemble of retrieval-and
  generation-based dialog systems.
\newblock {\em arXiv preprint arXiv:1610.07149}.

\bibitem[\protect\citeauthoryear{Sordoni \bgroup et al\mbox.\egroup
  }{2015}]{sordoni2015neural}
Sordoni, A.; Galley, M.; Auli, M.; Brockett, C.; Ji, Y.; Mitchell, M.; Nie,
  J.-Y.; Gao, J.; and Dolan, B.
\newblock 2015.
\newblock A neural network approach to context-sensitive generation of
  conversational responses.
\newblock In {\em NAACL},  196--205.

\bibitem[\protect\citeauthoryear{Srivastava \bgroup et al\mbox.\egroup
  }{2014}]{srivastava2014dropout}
Srivastava, N.; Hinton, G.; Krizhevsky, A.; Sutskever, I.; and Salakhutdinov,
  R.
\newblock 2014.
\newblock Dropout: a simple way to prevent neural networks from overfitting.
\newblock {\em The Journal of Machine Learning Research} 15(1):1929--1958.

\bibitem[\protect\citeauthoryear{Vinyals and Le}{2015}]{vinyals2015neural}
Vinyals, O., and Le, Q.
\newblock 2015.
\newblock A neural conversational model.
\newblock In {\em ICML (Deep Learning Workshop)}.

\bibitem[\protect\citeauthoryear{Wagner and Fischer}{1974}]{wagner1974string}
Wagner, R.~A., and Fischer, M.~J.
\newblock 1974.
\newblock The string-to-string correction problem.
\newblock {\em Journal of the ACM (JACM)} 21(1):168--173.

\bibitem[\protect\citeauthoryear{Weizenbaum}{1966}]{weizenbaum1966eliza}
Weizenbaum, J.
\newblock 1966.
\newblock Eliza—a computer program for the study of natural language
  communication between man and machine.
\newblock {\em Communications of the ACM} 9(1):36--45.

\bibitem[\protect\citeauthoryear{Weston, Dinan, and
  Miller}{2018}]{weston2018retrieve}
Weston, J.; Dinan, E.; and Miller, A.~H.
\newblock 2018.
\newblock Retrieve and refine: Improved sequence generation models for
  dialogue.
\newblock {\em arXiv preprint arXiv:1808.04776}.

\bibitem[\protect\citeauthoryear{Williams}{1992}]{williams1992simple}
Williams, R.~J.
\newblock 1992.
\newblock Simple statistical gradient-following algorithms for connectionist
  reinforcement learning.
\newblock {\em Machine learning} 8(3-4):229--256.

\bibitem[\protect\citeauthoryear{Wu \bgroup et al\mbox.\egroup
  }{2018}]{wu2018response}
Wu, Y.; Wei, F.; Huang, S.; Li, Z.; and Zhou, M.
\newblock 2018.
\newblock Response generation by context-aware prototype editing.
\newblock {\em arXiv preprint arXiv:1806.07042}.

\bibitem[\protect\citeauthoryear{Xing \bgroup et al\mbox.\egroup
  }{2017}]{xing2017topic}
Xing, C.; Wu, W.; Wu, Y.; Liu, J.; Huang, Y.; Zhou, M.; and Ma, W.-Y.
\newblock 2017.
\newblock Topic aware neural response generation.
\newblock In {\em AAAI},  3351--3357.

\bibitem[\protect\citeauthoryear{Xu \bgroup et al\mbox.\egroup
  }{2018}]{unpaired-sentiment-translation}
Xu, J.; Sun, X.; Zeng, Q.; Ren, X.; Zhang, X.; Wang, H.; and Li, W.
\newblock 2018.
\newblock Unpaired sentiment-to-sentiment translation: A cycled reinforcement
  learning approach.
\newblock In {\em ACL},  675–686.

\bibitem[\protect\citeauthoryear{Zhao, Lee, and
  Eskenazi}{2018}]{zhao2018unsupervised}
Zhao, T.; Lee, K.; and Eskenazi, M.
\newblock 2018.
\newblock Unsupervised discrete sentence representation learning for
  interpretable neural dialog generation.
\newblock In {\em ACL},  1098--1107.

\bibitem[\protect\citeauthoryear{Zhao, Zhao, and
  Eskenazi}{2017}]{zhao2017learning}
Zhao, T.; Zhao, R.; and Eskenazi, M.
\newblock 2017.
\newblock Learning discourse-level diversity for neural dialog models using
  conditional variational autoencoders.
\newblock In {\em ACL},  654--664.

\bibitem[\protect\citeauthoryear{Zhou \bgroup et al\mbox.\egroup
  }{2018}]{zhou2018commonsense}
Zhou, H.; Young, T.; Huang, M.; Zhao, H.; Xu, J.; and Zhu, X.
\newblock 2018.
\newblock Commonsense knowledge aware conversation generation with graph
  attention.
\newblock In {\em IJCAI},  4623--4629.

\end{thebibliography}
	\bibliographystyle{aaai}
\end{document}